\documentclass[11pt,a4paper]{article}
\usepackage[hyperref]{acl2019}
\usepackage{times}
\usepackage{latexsym}

\usepackage{url}

\usepackage{booktabs}
\usepackage{url}
\usepackage{amsmath}
\usepackage{float}
\usepackage{graphicx}
\usepackage{listings}
\usepackage{multirow}
\usepackage{makecell}
\usepackage{todonotes}
\usepackage{subcaption}

\aclfinalcopy % Uncomment this line for the final submission
\newif\ifcomments
\commentstrue 
\ifcomments
    \newcommand{\pinaforecomment}[3]{\colorbox{#1}{\parbox{.95\linewidth}{#2: #3}}}
\else
    \newcommand{\pinaforecomment}[3]{}
\fi

\newcommand{\spider}{\textsc{Spider}}

\title{Grammar-based Neural Text-to-SQL Generation}

\author{Kevin Lin$^{1}$ ~~~~ Ben Bogin$^{2}$ ~~~~ Mark Neumann $^{1}$\\
~~~~~ \textbf{Jonathan Berant}$^{1,2}$ ~~~~~ \textbf{Matt Gardner}$^{1}$  \\
\mbox{}\\
$^1$Allen Institute for Artificial Intelligence \\
$^2$School of Computer Science, Tel-Aviv University \\
\small{\texttt{\{kevinl, markn, mattg\}@allenai.org} \hspace{0.1cm} \small{\texttt{\{ben.bogin, joberant\}@cs.tau.ac.il}}}}

\date{}

\begin{document}
\maketitle
\begin{abstract}
  The sequence-to-sequence paradigm employed by neural text-to-SQL models typically performs token-level decoding and does not consider generating SQL hierarchically from a grammar. Grammar-based decoding has shown significant improvements for other semantic parsing tasks, but SQL and other general programming languages have complexities not present in logical formalisms that make writing hierarchical grammars difficult.  We introduce techniques to handle these complexities, showing how to construct a schema-dependent grammar with minimal over-generation. We analyze these techniques on ATIS and \spider{}, two challenging text-to-SQL datasets, demonstrating that they yield 14--18\% relative reductions in error.
\end{abstract}

\section{Introduction}
\label{sec:intro}
Natural language interfaces to databases (NLIDB), the task of mapping natural language utterances to SQL queries, has been of interest to both the database and natural language processing communities, as effective NLIDB  would allow people of all technical backgrounds to access information stored in relational databases.

Recent text-to-SQL models typically take a standard sequence-to-sequence modeling approach, encoding a sequence of natural language tokens and then decoding a sequence of SQL tokens, possibly constrained by the table schema or a SQL grammar in some way \citep{Iyer2017LearningAN,yu2018typesql,yu2018syntaxsqlnet}.  However, work in the (closely related) semantic parsing literature has shown that hierarchical, grammar-based decoding, where the output of the model changes from a sequence of tokens to a sequence of productions rules from the grammar, is often more effective \cite{rabinovich2017abstract,krishnamurthy2017neural,yin2017syntactic}.

Applying grammar-based decoding to general programming languages such as SQL is very challenging.  Constructing a grammar that constrains the outputs correctly such that it cannot generate invalid programs (``over-generate'') is difficult, as the abstract syntax trees \citep[ASTs;][]{aho1986compilers} used by the languages' compilers \footnote{The same also applies for interpreters; we use the term compilers in this work to simplify the discussion.} are not sufficiently constraining.  There are trade-offs between manual effort in constructing a tight grammar, the complexity and depth of the grammar, and the learnability of the grammar to a model.  These languages often define typed variables (e.g., table aliases in SQL), which mean they are not context-free, requiring more complex mechanisms to handle and making it difficult to construct a grammar that completely removes over-generation.  There are often classes or schemas that need to be respected when generating (e.g., table columns like \texttt{city.city\_name}), requiring the grammar to depend on the schema of the database being queried.  With SQL, this can be taken one step further, constraining (or at least encouraging) comparisons on table columns to be values in that column (e.g., \texttt{WHERE city.city\_name = "New York"}).

In this work we develop a grammar that covers more than 98\% of instances with minimal over-generation in two popular datasets: ATIS~\citep{hemphill1990atis}, a dataset of contextual interactions with a flight database, and \spider{}~\citep{yu2018spider}, a dataset focused on complex SQL queries over a variety of schemas, many of which are unseen at test time.  We show how to modify grammar-based semantic parsers to use this grammar, and discuss how the common practice of identifier anonymization in SQL queries applies to grammar-based decoding.  Interestingly, prior grammar-based parsers have their own linking mechanism which serves largely the same purpose as identifier anonymization, and we show that these two mechanisms are complementary to each other.  Finally, we note that context-sensitive grammar constraints are easily handled inside the decoder, allowing us to use a relatively simple context-free grammar and impose further constraints (e.g., on the production of joins in SQL) at run-time (both during training and inference).

We apply these contributions to models for ATIS and \spider{}, demonstrating the effectiveness of grammar-based decoding for text-to-SQL tasks.  Our model achieves 73.7\% denotation accuracy on the resplit, contextual ATIS task \cite{Suhr2018LearningTM}, a 4.5\% absolute improvement over the prior best result, and 33.8\% accuracy on the database split of \spider{}, a 14.1\% absolute improvement over the best prior work with the same supervision.

\section{SQL Grammar}
\label{sec:grammar}
In this section we discuss several important considerations when designing a grammar for a general programming language like SQL, and we present the grammar that we use in our experiments.

When doing grammar-based decoding on a programming language, one obvious potential starting place, which has been used repeatedly in prior work, is to directly use a compiler's grammar and the ASTs it produces \cite{yin2017syntactic,Iyer2018MappingLT}.  This approach, while simple and intuitive, has several drawbacks.  First, these grammars are written to \emph{recognize} and \emph{parse} presumed-valid programs, and further checking is done by the compiler after the ASTs are produced.  This means that using the compiler's grammar for grammar-based decoding will significantly \emph{over-generate} programs, still requiring a semantic parser to learn which of the possible programs that it can produce are actually valid programs.  Second, these grammars are also typically \emph{very deep}, with many intermediate non-terminals and unary productions that lead to very long derivations for simple programs.  It is easier for a semantic parser to learn to produce shorter derivations, so a shallower grammar would be preferable.

The main issue that leads a compiler's grammar to over-generate in a semantic parser is that a programming language is not context free, while the compiler's grammar for it generally is.  The context-sensitive parts of a programming language revolve around variables, their definitions, and their use.  A variable can have user- or schema-defined types, which restrict the identifiers that are validly used in conjunction with it. 
For example, a class in python would only have a limited set of member variables and functions, and a SQL identifier referring to a table in a database, such as \texttt{city}, only has a limited set of column identifiers that can be used with it, such as \texttt{city.city\_name}.

\begin{figure}
  \centering
  \includegraphics[width=\columnwidth]{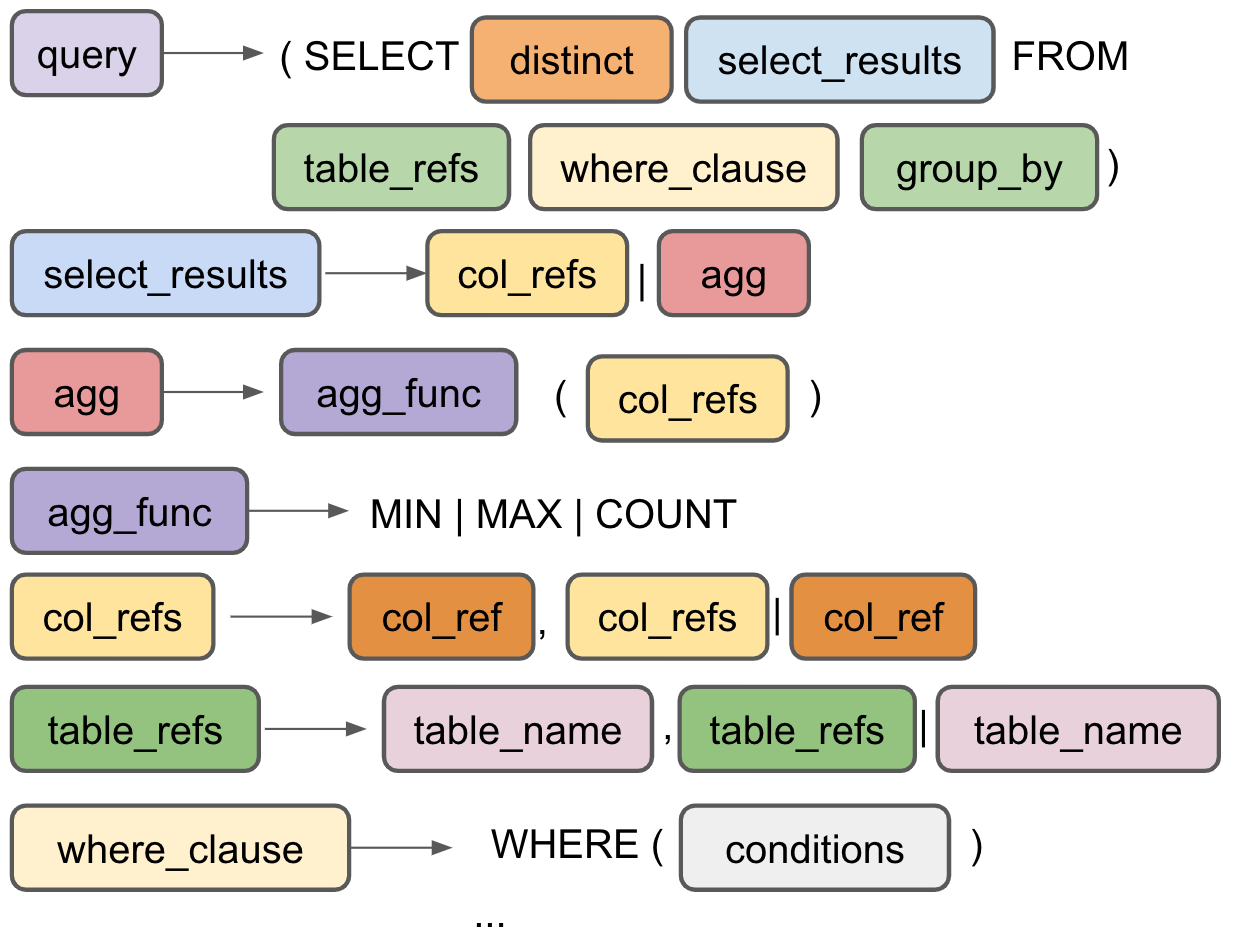}
  \caption{The base SQL grammar before augmentation with schema specific and utterance specific rules. The non-terminals are shown in boxes.
  }
  \label{fig:atis_grammar}
\end{figure}

We address these issues for SQL by designing shallow parsing expression grammars \citep{ford2004parsing} that capture the minimum amount of SQL necessary to cover most of the examples in a given dataset.
Limiting the SQL covered to only what is necessary allows the grammar to be more compact, which aids the learnability of the grammar for the semantic parser.  Unfortunately, this means that the grammars we write are dataset-specific, though we share a common base that needs only minimal modification for a new dataset. A simplified base grammar is shown in Figure~\ref{fig:atis_grammar}. \footnote{The full grammar and code for reproducing the experiments are in AllenNLP \cite{gardner2018allennlp}}

In order to handle the context-sensitive components of SQL, we use two approaches.  First, we note that some amount of context sensitivity can be handled by adding additional non-terminals to a context-free grammar (c.f. \citet{Petrov2006LearningAC}), and we use this approach to ensure consistency of table, column, and value references.  Second, for more complex context sensitivity, such as ensuring that joined tables in a SQL query share a common foreign key, we use runtime constraints on production rule expansion (during decoding at both training and inference time) to ensure that only valid programs are allowed (c.f. \citet{Liang2017NeuralSM}).

\begin{figure}[t]
  \centering
  \includegraphics[width=\columnwidth]{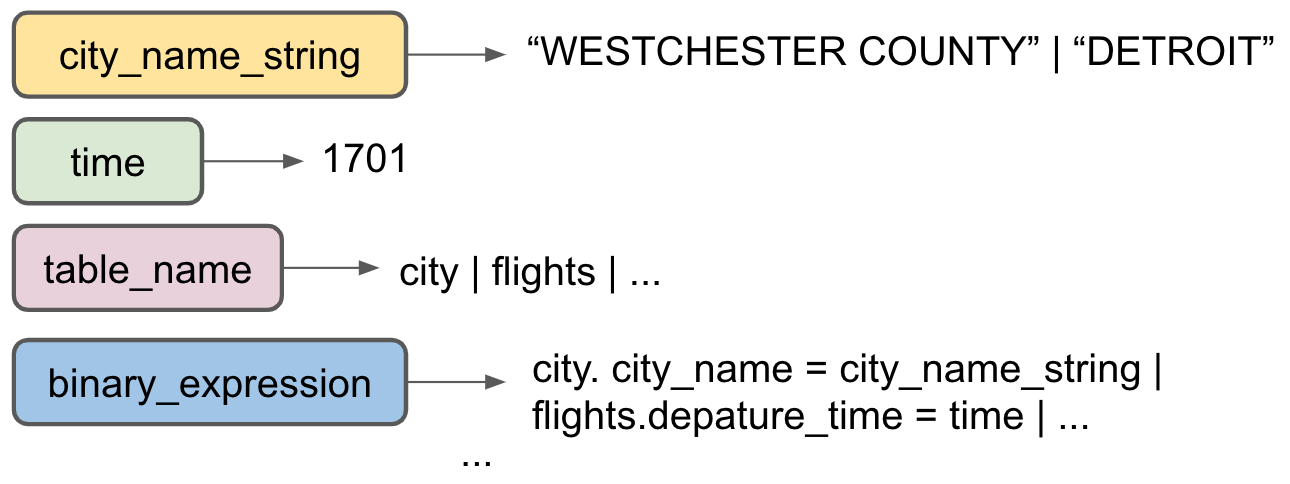}
  \vspace{-2mm}
  \caption{Example of additional rules added to the base SQL grammar based on database schema and entities in utterance if the entities \texttt{WESTCHESTER COUNTY}, \texttt{DETROIT}, and \texttt{1701} are detected in the utterance}
  \label{fig:additional_grammar_rules}
  \vspace{-5mm}
\end{figure}

\textbf{Adding schema non-terminals:} In the base grammar, the \texttt{table\_name} and \texttt{col\_ref} non-terminals are left undefined. For each example, both during training and inference, we examine the database schema associated with that example and automatically add grammar rules for these non-terminals.  All tables in the database have their names added as valid productions of the \texttt{table\_name} non-terminal, and each column in each table gets a production for the \texttt{col\_ref} non-terminal that generates the table and column name together (e.g., \texttt{city.city\_name}).  We further only allow comparisons to table columns with values that occur in that column.  For example, in a \texttt{WHERE} clause, we only allow statements such as \texttt{city.city\_name = VALUE} where \texttt{VALUE} is actually a value in the \texttt{city.city\_name} column in the database.  We accomplish this by modifying the \texttt{biexpr} non-terminal to have one possible production for each table column, making use of a new non-terminal for values in that column.  An example of each of these kinds of rules is shown in Figure~\ref{fig:additional_grammar_rules}.  Note that an compiler's grammar would allow arbitrary identifiers in these conditions; in order to properly constrain the productions allowed by the semantic parser in a given context, we need to add these schema-dependent production rules.

The binary comparison rule mentioned above, restricting column comparisons to only accept values in the corresponding column, is occasionally too strict.  If our input utterance mentions ``flights before 5:01pm'', we want to be able to have a clause like \texttt{WHERE flight.departure\_time < 1701}.  In order to handle cases like this, we additionally examine the input utterance and dynamically add rules to the grammar based on values seen there.  These are largely based on heuristic detection of numbers and times in the input.  This is also shown in Figure \ref{fig:additional_grammar_rules}.

Both of these mechanisms for dynamically producing production rules, either from the database schema or from the input utterance, can generate rules at test time that were never seen during training.  In order to handle this, we distinguish between \emph{global} rules that come from the base grammar, and \emph{linked} rules that are dynamically generated.  These two kinds of rules will be parameterized differently in the model (\S\ref{sec:model}), so that the model can handle unseen rules at test time.

\textbf{Run-time Grammar Constraints:}  For datasets that involve joins we apply additional constraints at run-time. To start off, we keep track of two sets of tables: used tables, $U$, and required tables, $R$. When a \emph{table} is \texttt{SELECT}ed or \texttt{JOIN}ed, it is added to the set of used tables, and when a \emph{column} is \texttt{SELECT}ed, the table that it belongs to is added to a set of required tables. . First, when generating \texttt{WHERE}, \texttt{ORDER BY}, \texttt{GROUP BY}, and \texttt{JOIN} conditions, we eliminate rules that generate columns that are not from the set of used tables. Second, when predicting the last join, if there exists a table $t$ in $R$ not in $U$, we remove all rules that do not join $t$. Third, we constrain the number of joins using the used tables and required tables. If $|R| - |U| > 1$, then there must be more joins so we remove all rules that stop joins. If $|R| - |U| \leq 1$, then we do not allow rules that generate more than one join, since we are assuming no self-joins.

\textbf{Other Considerations:} Many current text-to-SQL datasets in the NLP community make liberal use of table aliases when they are not strictly necessary.  These aliases give traditional sequence-to-sequence models some consistency when predicting output tokens, but unnecessarily complicate the grammar in grammar-based decoding.  It makes the grammar deeper, and requires the parser to keep track of additional identifiers that are hard to model.  Accordingly, we simply undo the table alias normalization that has been done in these datasets before training our model, and add it back in during post-processing of our predicted queries if the dataset requires it.  Table aliases \emph{are} sometimes required in complex SQL programs, but these are very rare in current datasets, and we do not currently handle them.

\textbf{Linearizing a syntax tree:} Given this dynamically generated grammar for a given example, during training we parse the input SQL into an AST.  Following \citet{krishnamurthy2017neural}, we then linearize this tree depth-first, left-to-right, to get a sequence of production rules for the parser to learn to generate.  During decoding, the grammar (along with runtime constraints) is used to constrain the production rules available to the model at each timestep.
An example query derivation in this grammar can be seen in Figure~\ref{fig:grammar}.

\begin{figure}[h]
    \centering
    \begin{subfigure}{0.48\textwidth}
    \begin{lstlisting}[language=SQL, keywordstyle=\color{blue}, basicstyle=\fontfamily{cmtt}\small,columns=fullflexible,frame=bt, breaklines=true]
SELECT FLIGHT . COST,
FROM FLIGHT
WHERE FLIGHT .  FLIGHT_TIME = 
    (SELECT MIN(FLIGHT . FLIGHT_TIME)
     FROM FLIGHT);
    \end{lstlisting}
    \caption{Gold SQL label}
    \end{subfigure}
    
    \begin{subfigure}{0.48\textwidth}
    \begin{lstlisting}[basicstyle=\fontfamily{cmtt}\small,columns=fullflexible,frame=bt,  breaklines=true]
statement -> [query, ";"]
query -> ["(", "SELECT", distinct, select_results, "FROM", table_refs, where_clause, ")"]
distinct -> ""
select_results -> [col_refs]
...
    \end{lstlisting}
    \caption{Gold derivation}
    \end{subfigure}
    \caption{An example of how gold SQL queries are transformed into gold derivations for model supervision. Derivations are formed from a depth-first traversal of the AST from the parsed statement.}
    \label{fig:grammar}
\end{figure}

\section{Model}
\label{sec:model}

\begin{figure*}[t]
    \includegraphics[width=\textwidth]{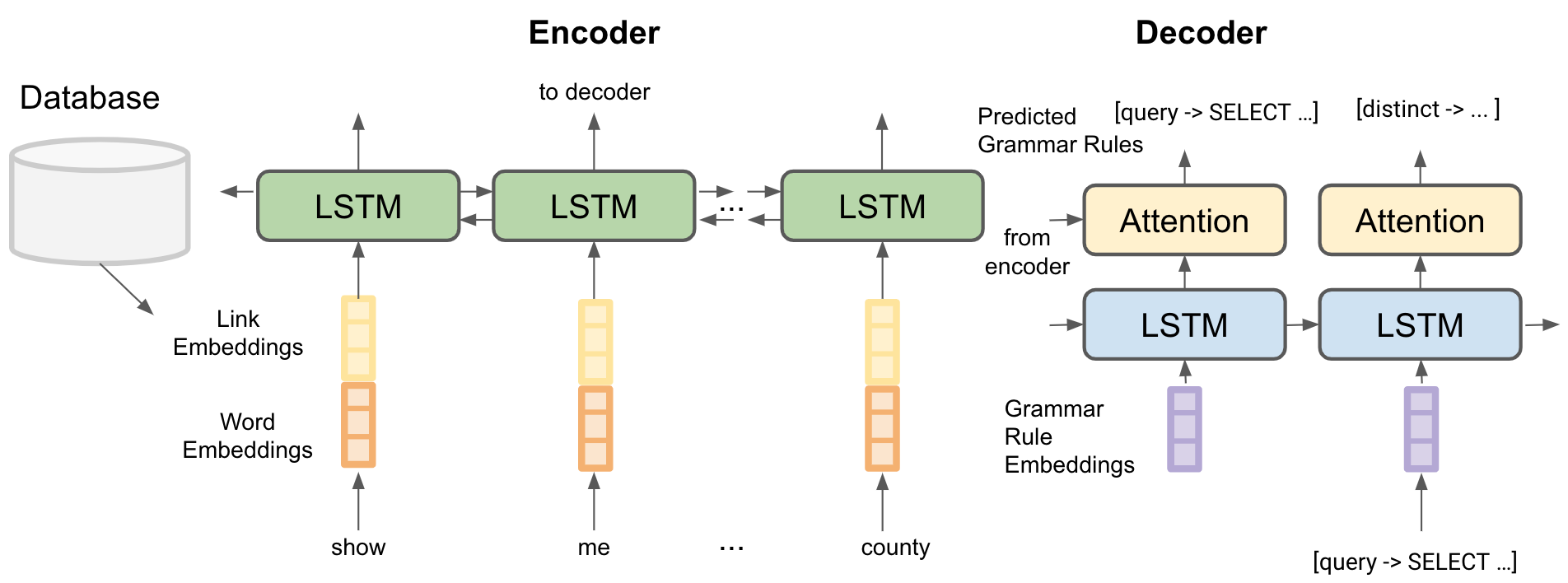}
        \caption{Overview of our type-constrained semantic-parser based on \citet{krishnamurthy2017neural}. The encoder links input tokens to database values and generates link embeddings which, along with word embeddings, are fed into a bidirectional LSTM. The decoder predicts a sequence of SQL grammar rules that generate a SQL query.}
    \label{fig:model}
\end{figure*}

To translate natural language utterances to SQL statements, we pair our grammar from Section~\ref{sec:grammar} with a semantic parsing model that closely follows that of \citet{krishnamurthy2017neural}.\footnote{The main differences with prior work are in how and when we compute linking scores, and in the identifier anonymization.}  Our model takes as input an utterance, a database, and an utterance-specific grammar, and outputs a sequence of production rules that sequentially build up an AST for a SQL program.  The model distinguishes between two kinds of production rules: (1) \emph{global} rules that come from the base grammar and are shared across utterances, and (2) \emph{linked} rules that are utterance-specific and might be unseen at test time.  The base grammar rules typically determine the \emph{structure} of the SQL statement, while the utterance-specific rules perform \emph{linking} of words in the utterance to identifiers in the database (such as table names, columns names, and column values).

\textbf{Notation}: The utterance is denoted as a sequence of tokens $[u_1, \ldots, u_n]$. Identifiers in the database that may be unseen at test time, such as the name of a city or an airport code in ATIS, or table and column names in \spider{}, are denoted as $e$, and the whole set of identifiers is denoted as $E$.  The production rule that generates a particular identifier is denoted as $l_e$.

\textbf{Identifier Linking}
We use simple string matching heuristics to link words or phrases in the input utterance to identifiers in the database.\footnote{This heuristic string matching could easily be replaced by a learned function, and this was done in prior work that focused on WikiTableQuestions~\citep{krishnamurthy2017neural}, but we found that to be unnecessary for these text-to-SQL datasets.} For example, if ``Boston`` appears in the utterance, then it should be linked to the linked rules that produce the relevant identifiers, such as \texttt{city\_name\_string -> "BOSTON"} and \texttt{city\_code -> "BOS"}. First, we generate a linking score between the utterance token, $u_i$ and each identifier: 
\begin{align*}
    s(e, u_i) &=
    \begin{cases}
      1, & \text{if}\ u_i \text{ heuristically triggers } e \\
      0, & \text{otherwise}
    \end{cases} 
\end{align*}
We use this linking score in both the encoder and the decoder. In the encoder, we generate a \emph{link embedding} for each token in the utterance that represents what database values it is linked to. For each linked rule that generates a database identifier $e$, we generate a type vector $v_{\tau(e)}$ based on the non-terminal type of the identifier. This allows the model to handle unseen identifiers at test time. The link embedding is then computed as $l_i = \sum_{e \in E} s(e, u_i) \tanh{v_{\tau(e)}}$. The decoder section describes the use of the linking in decoding.

\textbf{Encoder:} The encoder is a bi-directional LSTM \cite{hochreiter1997long} that takes as input a concatenation of a learned word vector and the link embedding for each token. To incorporate the history of the interaction in ATIS, we concatenate the previous $n$ utterances delimited with special tokens. The model is able to access the previous utterances but not the previous queries.

\textbf{Decoder:} The decoder is an LSTM with attention on the input utterance that predicts production rules from the grammar described in Section \ref{sec:grammar}. At each step, the decoder iteratively builds up the SQL query by applying a grammar rule to the leftmost non-terminal in the AST. The production rules associated with any particular non-terminal could either be global rules, linked rules, or both. Global rules are parameterized with an embedding, and the model assigns logits to these rules using a multilayer perceptron. Linked rules are parameterized using the decoder's attention over the input utterance and the linking scores mentioned earlier. At step $j$, the decoder computes an attention $a_j$ over the input utterance and then computes logits for linked rules as $s_j(l_e) = \sum_{i} s(e, i) a_{ji}$.  Logits for all rules are jointly normalized with a softmax to produce a distribution over the available production rules at each decoding step.  We note that this parameterization of linked rules through the attention mechanism is a key difference from traditional sequence-to-sequence models.  It is similar to a copy mechanism \citep{Gu2016IncorporatingCM}, though we are ``copying'' production rules that are \emph{linked} to utterance tokens, not the utterance tokens themselves.

\textbf{Identifier anonymization}, which has long been done in text-to-SQL models, is the process of taking database identifiers that appear in both the question and SQL query and replacing them with dummy variables, to simplify the prediction task. For example, the utterance ``what flights go from boston to orlando'' would be preprocessed to be ``what flights go from \texttt{CITY\_NAME\_0} to \texttt{CITY\_NAME\_1}''. This anonymization has some of the same goals as our identifier linking---enabling prediction of identifiers not seen during training---but it also simplifies the \emph{encoder's vocabulary}, because all city names get removed from the vocabulary and replaced with the dummy variable.  In our model, we experiment with both anonymization and linking at the same time, treating the dummy variables as linked production rules.  Importantly, however, we do the anonymization using our linking heuristics \emph{only}, not looking at the SQL query, so our evaluation is equivalent to a non-anonymized setting.

\textbf{Training} The model is given access to utterances paired with one or more corresponding SQL queries. The SQL queries are parsed into their derivations, which are used as supervision for the model. The model is then trained to maximize the log-likelihood of the labeled query. If there are multiple programs we train on only the one with the shortest derivation.

\section{Experiments}
\label{sec:experiments}

\begin{table}[ht]
\begin{tabular*}{0.48\textwidth}{lp{0.6cm}p{0.6cm}p{0.6cm}p{0.6cm}}
\toprule
                  & \multicolumn{2}{c}{\bf Development} & \multicolumn{2}{c}{\bf Test} \\

    & \textbf{Q}        & \textbf{D}       & \textbf{Q}    & \textbf{D}   \\ \midrule
                  \citet{Suhr2018LearningTM}                  & 37.5         & 62.5             & 43.6     & 69.2          \\ \midrule 

Ours                     & \textbf{39.1}         & \textbf{65.8}             & \textbf{44.1}       & \textbf{73.7}          \\ \bottomrule
\end{tabular*}
\caption{Comparison of our model with the best prior work on the ATIS dataset. \textbf{Q} and \textbf{D} correspond to exact query accuracy and denotation accuracy. The main difference between the models is that ours uses grammar-based decoding while \citet{Suhr2018LearningTM}'s is token-based.}
\label{table:prior_work}
\end{table}

\begin{table}[ht]
\centering
\begin{tabular}{lrr}
\toprule 
                 & \textbf{Development}  & \textbf{Test} \\  \midrule

\citet{yu2018syntaxsqlnet}       &   18.9         &      19.7        \\ \midrule 

 Ours                    &      \textbf{34.8}              &    \textbf{33.8}          \\ \bottomrule
\end{tabular}
\caption{Comparison of our model with the best prior work on the \spider{} dataset, with the exact component matching accuracy. }
\label{table:spider}
\end{table}

\subsection{Datasets}
We evaluate on two datasets, the ATIS flight planning dataset \citep{hemphill1990atis} and the \spider{} dataset \citep{yu2018spider}.

\textbf{ATIS}: We use \citet{Suhr2018LearningTM}'s data re-split to avoid scenario bias and to make use of their pre-processing to identify times with UWTime \citep{Lee2014ContextdependentSP}. The dataset consists of 1148/380/130
train/dev/test interactions. There is an average of 7 utterances per interaction.  We use \citet{Suhr2018LearningTM}'s model as our baseline, as it uses the same dataset, preprocessing, and supervision, but with token-based decoding.

For evaluation, we use exact query match accuracy and denotation accuracy. Query match accuracy is the percentage of queries that have the same sequence of SQL tokens as the reference query. Denotation accuracy is the percentage of queries that execute to the same table as the gold query, where credit is not given to queries that do not execute. Denotation accuracy is a particularly important evaluation metric when considering SQL as a target language, as the ordering of various clauses do not affect query execution \citep{xu2017sqlnet}.

\textbf{\spider{}}: The key difference between Spider and other text-to-SQL datasets is that databases not seen in training can appear in the test set. In the database split, databases are split randomly into 146/20/40 train/dev/test databases. To do well on the database split, the model needs to learn to compose various SQL operators and generalize to new schemas, as all databases will be unseen at test time.
To compare with prior work, we report the exact component matching score. The predicted query is decomposed by the \texttt{SELECT}, \texttt{WHERE}, \texttt{GROUP BY}, \texttt{ORDER BY} and \texttt{KEYWORDS}. Each component in the predicted query and the ground truth are then decomposed into subcomponents and checked if the sets of the components match exactly. The predicted query is correct when all components match.

\subsection{Implementation Details} 
The two datasets we experiment with were designed to test different aspects of the text-to-SQL task, and thus we include different parts of the model in each dataset to address them. Our model for the ATIS dataset includes identifier anonymization---since this dataset is evaluated on execution accuracy linking tokens in the utterance to database values is extremely important. The ATIS model does not include the run-time constraints as there are no joins or table aliases in this dataset. Conversely, \spider{} already anonymizes database values but has many joins, so our model does not have identifier anonymization and database value generation but does include run-time constraints.

We use the sparse Adam optimizer with a learning rate of 0.001 \cite{kingma2014adam}. We use a batch size of 32 and initial patience of 10 epochs. We use accuracy on the dev set as a metric for early stopping and hyperparameter tuning. We use uniform Xavier initialization for the weights of the LSTM and zero vectors for the biases \cite{glorot2010understanding}. The word embeddings and identifier type embeddings are both of size 400 are not pretrained. The encoder and decoder both contain 1 layer with hidden size 800. We apply dropout with probability of 0.5 after the encoder. During training, we train on instances with derivation less than 300 steps and during inference we limit the decoder to 300 generation steps. For incorporating context on ATIS, we allow the model to see the past 3 utterance as context. During evaluation, we use beam search with a beam size of 10.

\subsection{Results}
Table \ref{table:prior_work} shows a comparison of our model against the best prior published result on the context-dependent ATIS dataset.  Our model, which includes identifier linking, link embeddings and type-constraints, yields a 4.5\% improvement in denotation accuracy over prior work.  Table \ref{table:spider} shows a comparison of our model with previous work on the \spider{} dataset, and shows that it yields a 14.1\% increase in exact component matching compared to the best previously published result.\footnote{\citet{yu2018syntaxsqlnet} also present a model that gets 27.2\% accuracy, but it uses additional manual annotations. Table \ref{table:spider} shows what is included on the official \spider{} leaderboard.}

\section{Discussion}
\label{sec:discussion}

\begin{table}
\centering
\begin{tabular}{lllll}
\toprule

\textbf{Linking} & \textbf{Link Embedding}   & \textbf{Anon.} & \textbf{Acc.}   \\ 
\midrule

Yes & No                   & No                   & 57.1            \\
Yes & Yes                  & No                & 60.4            \\ 
Yes & No                   & Yes              & 64.1            \\
No & Yes                  & Yes              &  60.6             \\
\midrule
Yes & Yes                  & Yes                   & \textbf{65.8}          \\
\bottomrule
\end{tabular}
\caption{Model ablations on ATIS, ablating the linked rules during decoding (making them global rules), the link embedding, and the identifier anonymization, showing denotation accuracy on the development set.}
\label{table:model-ablation}
\end{table}

Table \ref{table:model-ablation} presents ablations of various components of the model.  In the setting without identifier anonymization, the link embedding improves denotation accuracy by 3.3\%. This is due to the fact that identifiers need to be accounted for not only to generate values, but also to generate the correct query structure. Figure \ref{fig:example_query} shows that model \textit{with} link embeddings is able to use the type information to generate the correct query structure and values, even for identifiers that have low frequency in the dataset. In this case, even before the model generates the linked identifier, \texttt{fare\_basis . fare\_basis\_code = `F'}, the model has to generate the correct columns in the \texttt{SELECT} clause and table in the \texttt{FROM} clause. With the link embedding, the model correctly identifies that it needs to select from the \texttt{fare\_basis} table, while the model without the link embedding incorrectly selects the \texttt{class\_of\_service} table. % In the setting with identifier anonymization, the link embedding provides a smaller gain of 1.7\% in denotation accuracy, as anonymization and link embeddings provide similar (but not completely overlapping) information to model.

\begin{table}
\centering
\begin{tabular}{lll}
\toprule
   \textbf{Constrained Columns} & \textbf{Values} & \textbf{Acc.}   \\ 
\midrule

  No &          No    & 55.8            \\ 
 Yes & No             & 56.2            \\
\midrule
Yes & Yes            &  \textbf{65.8}            \\
\bottomrule
\end{tabular}
\caption{Grammar ablations on ATIS, ablating the column and value consistency constraints, showing denotation accuracy on the development set.}
\label{table:grammar-ablation-atis}
\end{table}

\begin{table}
\centering
\begin{tabular}{lll}
\toprule
   \textbf{Alias Pre.} & \textbf{Runtime Constraints} & \textbf{Acc.}   \\ 
\midrule

  No &          No    & 29.8            \\ 
 Yes & No             &  30.7             \\
 \midrule
Yes & Yes            &  \textbf{34.9}            \\
\bottomrule
\end{tabular}
\caption{Grammar abations on \spider{}, ablating the preprocessing for handling table aliases and run-time constraints, showing exact component accuracy on the development set.}
\label{table:grammar-ablation-spider}
\end{table}

Table \ref{table:grammar-ablation-atis} presents ablations of the schema-dependent grammar constraints on the ATIS dataset. We find that adding the constraint that columns appear with the table does not significantly improve performance. This could be due to the fact that the same tables are seen during training and test in ATIS, so associating tables with columns is not as challenging. However, removing the constraint on values decreases the denotation accuracy by 9.6\%, showing that generating the correct value in a \texttt{WHERE} clause is a central problem in this dataset. Table \ref{table:grammar-ablation-spider} shows that both table aliases and run-time constraints improve our model on for the \spider{} dataset.

\begin{figure}[t]
  \centering
    \begin{subfigure}[t]{0.48\textwidth}
    \begin{lstlisting}[language=SQL,keywordstyle=\color{blue},basicstyle=\fontfamily{cmtt}\small,columns=fullflexible,frame=bt, breaklines=true]
SELECT DISTINCT fare_basis . fare_basis_code,
                  fare_basis . booking_class,
                  fare_basis . class_type,
                  fare_basis . premium,
                  fare_basis . economy,
                  fare_basis . discounted,
                  fare_basis . night,
                  fare_basis . season,
                  fare_basis . basis_days
 FROM fare_basis
 WHERE fare_basis . fare_basis_code = 'F' ) ;
        \end{lstlisting}
    \caption{Model with link embeddings} \label{fig:link_embedding}
    \end{subfigure}

    \begin{subfigure}[t]{0.48\textwidth}
        \begin{lstlisting}[language=SQL,basicstyle=\fontfamily{cmtt}\small,keywordstyle=\color{blue},columns=fullflexible,frame=bt, breaklines=true]
SELECT DISTINCT class_of_service . booking_class,
                  class_of_service . rank,
                  class_of_service . class_description
FROM class_of_service
WHERE class_of_service . booking_class = 'F';
        \end{lstlisting}
    \caption{Model without link embeddings} \label{fig:no_link_embedding}
    \end{subfigure}
    
  \vspace{-2mm}
  \caption{Query generated with link embeddings \ref{fig:link_embedding} that matches the gold query and without link embedding \ref{fig:no_link_embedding} for the input utterance ``what is fare code f"}
  \label{fig:example_query}
  \vspace{-5mm}
\end{figure}

We also experimented with a production rule copy mechanism similar to that of \citet{Suhr2018LearningTM}. While copied production rules shorten the derivation and aid interpretability by showing which subtrees come from previous queries, we did not observe significant change in accuracy.

One final point that highlights the complexity of constructing grammars is that the ordering of recursive rules is important. For the \texttt{col\_refs} rule in Figure \ref{fig:grammar}, and similarly for \texttt{JOIN} clauses, switching between left or right branching can cause a several point difference in performance.  Automatically determining the optimal grammar is an interesting direction for future research.

\subsection{Error Analysis}
%While our grammar-based approach constrains the output to be syntactically valid SQL, it does not fully constrain the model to follow the database schema (e.g., the model may select a column from a table not in the \texttt{FROM} clause). However, this happens in only $0.01\%$ of queries produced by our model. In comparison, \citet{Suhr2018LearningTM} report approximately 7\% of the generated queries from their token level model are either not syntactically valid SQL or do not follow the database schema.

Since linked rules dealing with numbers and times are only added to the grammar based on the utterance text, the grammar will only parse a query correctly when \textit{all} identifiers in the query can be detected in the utterance. By manually inspecting the preprocessed utterances and gold SQL labels for ATIS, we found that UWTime identified incorrect dates in 27.6\% of the unparseable queries.
%Without the correct date, the model would subsequently generate the incorrect date or time. 
On the queries that can be parsed, our model performs substantially better, yielding 52\% query match accuracy and 80\% denotation accuracy, which suggests that improving datetime parsing could have a significant impact on performance.

In addition, we manually examined model output for 70 development set queries on the ATIS dataset. We found that 70\% of errors come from either linking or missing constraints. In particular, conflating references to airport tables and city tables was the cause of many errors, as references to cities and airports in the utterance are particularly ambiguous, resulting in poor linking using string heuristics. The remaining 30\% of errors stem from a variety of sources including difficulty in resolving anaphora, ambiguity in references to time, and selecting incorrect tables.

\section{Related Work}
\label{sec:related-work}
\textbf{Text-to-SQL}: Generating SQL queries from English queries has been a longstanding challenge that has interested both the database and NLP communities \citep{androutsopoulos1995natural}. More generally, semantic parsing into logical formalisms has been studied extensively in the NLP community \cite{data-geography-original,zettlemoyer2005learning,Liang2011LearningDC}. A relevant line of work in semantic parsing has been treating the problem as a sequence generation task by linearizing trees \citep{dong2016language, alvarez2016tree}.

\textbf{Datasets}:  We evaluate on the \spider{} and ATIS datasets, two datasets that present challenges not present in other text-to-SQL datasets. Spider is the most difficult in terms of query complexity and requires generalizing to unseen databases at test time \cite{yu2018spider}. ATIS requires handling context-dependent utterances and contains a large number of tables per database. Other well studied datasets include Restaurants \citep{data-restaurants-original}, Academic \citep{data-academic}  and WikiSQL \citep{zhongSeq2SQL2017}. There has recently been work in standardizing the many proposed text-to-SQL datasets \cite{FineganDollak2018ImprovingTE}.

\textbf{Grammar-based decoding}: Semantic parsers that output production rules from a grammar instead of directly outputting tokens have been studied for other formal languages such as $\lambda$-DCS and general purpose programming languages \citep{krishnamurthy2017neural, rabinovich2017abstract, yin2017syntactic}. Grammar-based methods have been explored by \citet{yin2018tranx} for WikiSQL. However, as noted by \citet{FineganDollak2018ImprovingTE}, WikiSQL is composed of relatively simple SQL queries, with over half of the queries of the form \texttt{(SELECT col AS result FROM table WHERE col = value)}, and can be parsed by just 4 grammar rules. The work most similar to ours is \citet{yu2018syntaxsqlnet}, which also exploits a SQL-specific grammar to constrain the output by structuring it as a set of recursive modules. However, they still output tokens instead of production rules, and have a more complex set of modules in their decoder. Our method considerably outperforms this work.

%Previous approaches to generating context-dependent queries for ATIS have used substantial intermediate structures as supervision \cite{miller1996fully}. Another approach has been parsing to $\lambda$-calculus \citep{zettlemoyer2009learning}. Neural sequence-to-sequence models with explicit references to previous generated SQL queries have also been proposed \citep{Suhr2018LearningTM}.

\textbf{Zero-shot semantic parsing}: One of the main challenges in \spider{} is handling databases at test time that were not seen during training, in a zero-shot setting.  Zero-shot semantic parsing has been studied before \citep{Herzig2018DecouplingSA,Lake2018GeneralizationWS}, with the best method using a complex two-step processes to decouple program structure from identifiers. Our grammar-based model, with separate handling for global rules and linked rules, naturally performs this decoupling without additional complexity.

\section{Conclusion}
\label{sec:conclusion}
We proposed a model that uses a dynamic schema-dependent SQL grammar to guide the decoding process and a deterministic entity linking module for the NLIDB task. Comparing to prior work, we show that decoding into a structured output with type constraints gives considerable improvements in performance, yielding a 4.5\% absolute increase in denotation accuracy and 14.1\% exact component matching over the best prior work on ATIS and \spider{} respectively. Our result suggests type information through link embedding or identifier anonymization and modeling context sensitivity is important for the task.

\bibliography{main}
\bibliographystyle{acl_natbib}

\end{document}